\documentclass{article}

\usepackage{arxiv}

\usepackage[utf8]{inputenc} % allow utf-8 input
\usepackage[T1]{fontenc}    % use 8-bit T1 fonts
\usepackage{color}
\usepackage{hyperref}
\hypersetup{colorlinks,citecolor=red,linkcolor=blue,urlcolor=cyan}
\usepackage{url}            % simple URL typesetting
\usepackage{booktabs}       % professional-quality tables
\usepackage{amsfonts}       % blackboard math symbols
\usepackage{nicefrac}       % compact symbols for 1/2, etc.
\usepackage{microtype}      % microtypography
\usepackage{cleveref}       % smart cross-referencing
\usepackage{graphicx}
\usepackage[numbers]{natbib}
\usepackage{doi}
\usepackage{listings}
\usepackage{siunitx}

\graphicspath{{./imgs/}}

\title{CLIPSE -- a minimalistic CLIP-based image search engine for research}

\newif\ifuniqueAffiliation
\uniqueAffiliationtrue

\author{ Steve Göring\hspace{1mm}\href{https://orcid.org/0000-0001-6810-6969}{\includegraphics[scale=0.06]{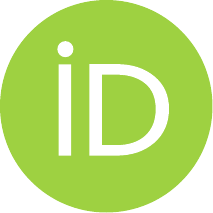}}\\
    Audiovisual Technology Group\\
    Technische Universität Ilmenau\\
    Germany \\
    \texttt{steve.goering@tu-ilmenau.de} \\
}

\renewcommand{\shorttitle}{\textit{arXiv} Template}

\hypersetup{
pdftitle={CLIPSE -- a minimalistic CLIP-based image search engine for research},
pdfsubject={image search engine},
pdfauthor={Steve Göring},
pdfkeywords={image search},
}

\begin{document}
\maketitle

\begin{abstract}
A brief overview of CLIPSE, a self-hosted image search engine with the main application of research, is provided.
In general, CLIPSE uses CLIP embeddings to process the images and also the text queries.
The overall framework is designed with simplicity to enable easy extension and usage.
Two benchmark scenarios are described and evaluated, covering indexing and querying time.
It is shown that CLIPSE is capable of handling smaller datasets; for larger datasets, a distributed approach with several instances should be considered.
\end{abstract}

% keywords can be removed

\section{Introduction}
Considering the increase of uploaded images per year, e.g., to photo sharing platforms (Flickr, Instagram, \ldots), or using AI text-to-image generators (DALL-E, Midjourney, \ldots), finding images to match a given text description is still an ongoing challenge.
Various image search engines, e.g., Google search, Bing search, or open-source engines (WISE~\cite{wise}, Image Search Engine~\cite{ise}), are available.
Recent advancements in machine learning move search engines from traditional text-based approaches using provided meta-data for images to use deep learning models.
For example WISE~\cite{wise} and Image Search Engine (ISE)~\cite{ise} both use vision-language models such as OpenCLIP~\cite{ilharco_gabriel_2021_5143773,cherti2023reproducible,Radford2021LearningTV,schuhmann2022laionb}.
The images for the search engine are transformed to one reduced embedding space, based on pre-training~\cite{Radford2021LearningTV}.
The pre-trained model was trained with (image, text) pairs so that even an arbitrary text can be transformed to this embedded space.
However, especially for research, small and easy to set up search engines considering own datasets are hard to find.
Usually image search engines rely on estimated meta-data, manually annotations (tags), or content analysis (computer vision/classification)~\cite{datta2008image}.
For this reason, a minimalistic search engine, called CLIPSE (CLIP-based image Search Engine), was developed.
CLIPSE is published as open-source software\footnote{\url{https://github.com/stg7/clipse}} and builds on Python~3, and HTML5 with CSS and Javascript.
The design follows a simplification approach; therefore, everything is kept as simple as possible.
This enables to perform a simple text-based image search for a given dataset, e.g., using command line or with a web-interface.

\section{Overview}
Similar to WISE and ISE, CLIPSE extracts for all images of a given dataset embeddings from OpenCLIP.
CLIPSE is a Python~3 application, and the setup is kept simple, UV~\cite{uv}(a fast and simple Python package manager) will automatically configure the dependencies (e.g. \texttt{pandas}, \texttt{flask}, \texttt{rich}, \texttt{tqdm}, and \texttt{open\_clip}).
CLIPSE further can process images using only CPU, therefore does not require a GPU.

\begin{figure}
\centering
\includegraphics[width=0.5\textwidth]{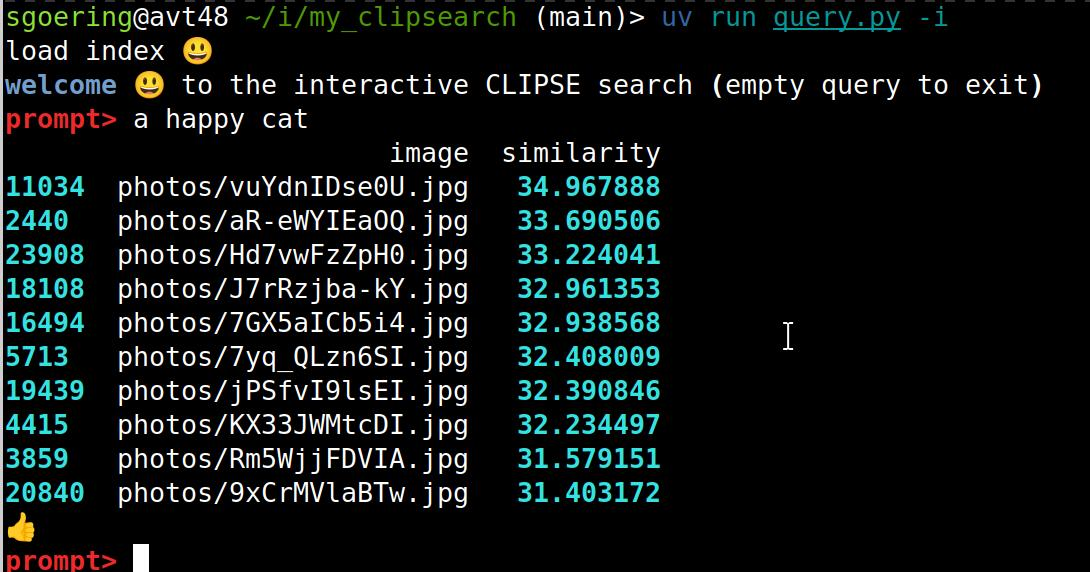}
\caption{CLIPSE command line interface, example query and result.}
\label{fig:cli}
\end{figure}

As a first step, the index must be created, here \lstinline[language={bash}]{build_index.py} can be called, with the folder of the images, it is recommended to resize them to e.g. $480\times480$ (or similar according to the aspect ratio) for faster processing.
The image is stored as \texttt{json} (for exploration) and compressed \lstinline[language={python}]{numpy.npz} (for faster loading).
After the index is created and stored, it can be accessed in two possible ways.
The first one is using the command-line \lstinline[language={bash}]{query.py} and produces colored output using the \texttt{rich} package.
A possible output is shown in Figure~\ref{fig:cli}.
\lstinline[language={bash}]{query.py} has several modes, e.g., for batch processing (no colors) or interactive sessions.

\begin{figure}
\centering
\includegraphics[width=0.85\textwidth]{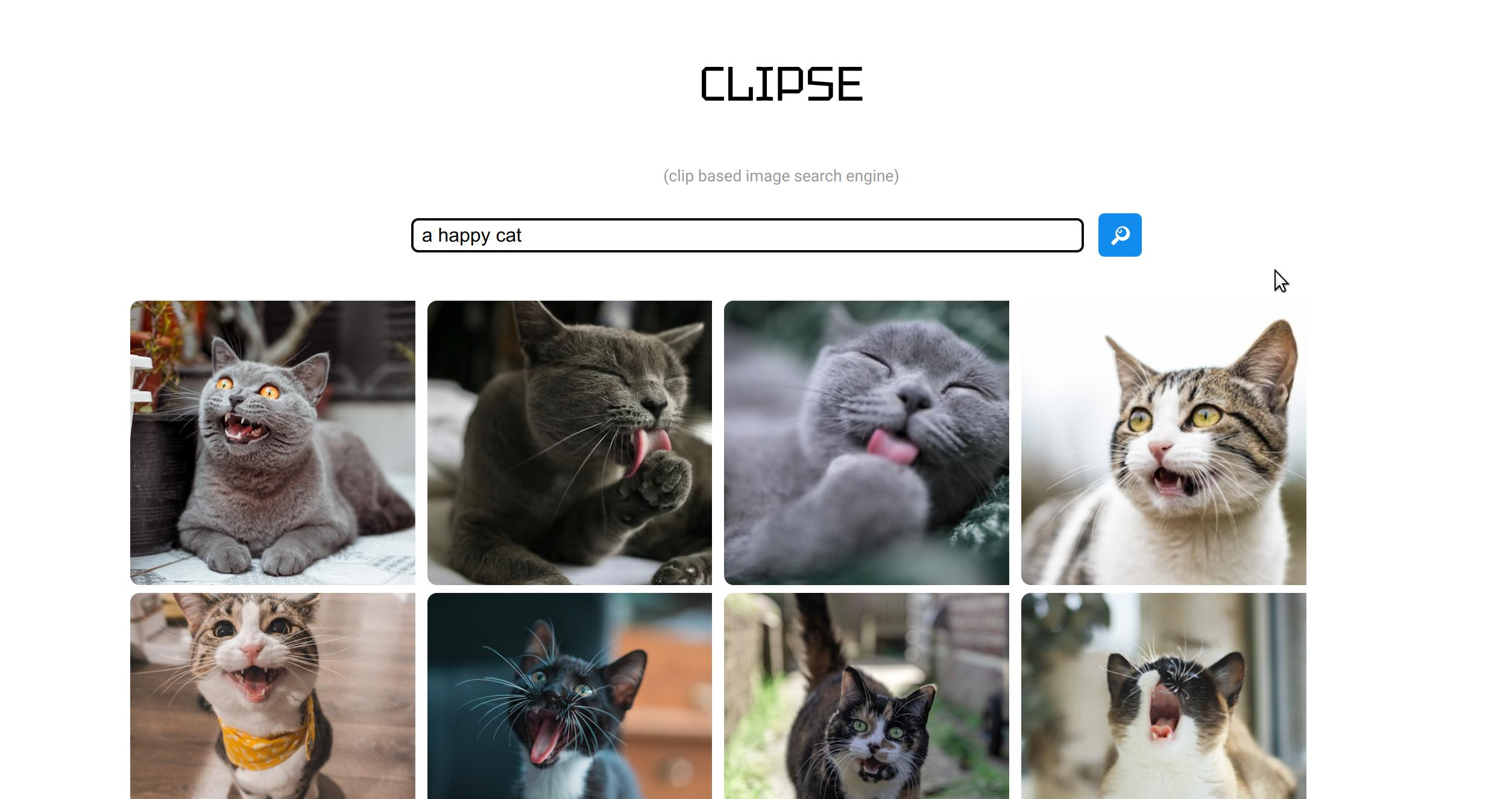}
\caption{CLIPSE web interface, example query and result.}
\label{fig:web}
\end{figure}

The second possible interfacing way is based on a web technology \lstinline[language={bash}]{server.py}.
An example for the web interface is shown in Figure~\ref{fig:web}.

The web interface is based on \texttt{flask} and one simple template that uses a minimal CSS framework (\texttt{MVP}) and two JavaScript functions (one to perform the server request to gather the search results, and another one for the pagination functionality).

In both interfacing cases, the text query is transformed using OpenCLIP to the same embedding space.
Afterward, the similarity of the text considering all images is calculated, based on the dot product.
The used dot product can be seen as an unnormalized cosine similarity measure.
Based on the similarity scores, the results are sorted and delivered as output to the corresponding interface.
Thus, the performance of CLIPSE depends on the used embedding model, which can be changed in the Python code.
Overall, to extend CLIPSE for very large datasets, several instances could run for subsets, and then a final aggregation can be performed of the results.

\section{Benchmark}
In the following section, small proof-of-concept benchmarks are performed, to evaluate the time to index a dataset and the time to perform queries.
All benchmarks are performed on an Intel NUC with an Intel Core i7-1185G7 CPU, 64 GB of main memory, a fast SSD, and no GPU acceleration.
PyTorch does parallel processing; however, the system performs indexing and querying sequentially.

The following test datasets are used:
\begin{itemize}
    \item \texttt{avt\_ai\_images}: 146 images based on text-to-image generation~\cite{goering2023ai,goering2023aiquality}
    \item \texttt{sample200}: 200 randomly selected images from the Unsplash-lite dataset~\cite{unsplash}
    \item \texttt{div2k}: all 900 images from the DIV2K dataset~\cite{agustsson2017ntire} (train and validation)
    \item \texttt{sophoappeal}: 1061 images from the SoPhoAppeal dataset~\cite{goering2023imageappeal}
    \item \texttt{avt\_image\_db}: 1133 images from~\cite{goering2019Intra}
    \item \texttt{unsplash-lite}: the full Unsplash-lite dataset~\cite{unsplash} consisting of $\approx 25k$ images
\end{itemize}

All the images have been resized before, to a resolution of $480\times W$ with $W$ adjusted to the aspect ratio of the image.
For all time measurements 32 repetitions have been performed.

\subsection{Time to build the index}

\begin{table}[htb!]
\centering
\caption{Measured time to index the datasets; values rounded to 3 decimals.}
\label{tbl:measured_time_to_index_the_datasets_}
\begin{tabular}{llrrr}
\toprule
dataset                  & \# images & average time [s] & std time [s] & avg time per image [s] \\
\midrule
\texttt{avt\_ai\_images} & 146       & 12.210            & 0.210         & 0.084 \\
\texttt{sample200}       & 200       & 14.554           & 0.087        & 0.073 \\
\texttt{div2k}           & 900       & 55.069           & 0.649        & 0.061 \\
\texttt{sophoappeal}     & 1061      & 61.458           & 0.111        & 0.058 \\
\texttt{avt\_image\_db}  & 1133      & 65.060            & 0.287        & 0.057 \\
\texttt{unsplash-lite}   & 24976     & 1411.237         & 10.144       & 0.057 \\
\bottomrule
\end{tabular}
\end{table}

The build index step consists of creating the embeddings, and then reading the index as a test and also to further convert the JSON file to a numpy file (for faster storing). The last step was deactivated for the measurements.
In Table~\ref{tbl:measured_time_to_index_the_datasets_} the time needed for the introduced differently sized datasets is listed.
Here, on average the time to index one images is nearly constant, e.g., $\leq 0.084\,s$.
For smaller datasets the overall overhead of starting the script and loading the dependencies has a larger impact, thus the time to index one image is slightly higher than for larger datasets.
Furthermore, it can be seen that the time to index is nearly linear growing based on the number of images.
It should be mentioned that no GPU acceleration and batch processing are used, both would enable a faster indexing time.
However, the overall time of $\approx23\,min$ for a dataset with $\approx 25k$ images is still acceptable.
A possible extension of the indexing process would be to include a dynamic increase of an already existing index, where new photo embeddings would be appended to the index.
For simplicity reasons this was dropped during the development.

\subsection{Time to process queries}
To evaluate the time for query processing, one test query is used, \texttt{query = "a cat exploring the dark night"}, other examples would result in similar time for the processing.
The query processing is done in three steps, first the index is loaded, then, as a second step, an embedded representation of the query is created, and as a last step, the similarities to all images are measured.
Thus, the overall size of the dataset has an influence on the response time for query processing.
In the next measurement, compare Table~\ref{tbl:measured_time_to_query_the_datasets_cold} a cold start is handled, thus the time to load the index and to perform the query is accumulated.
In the web interface, the index is usually loaded into main memory, thus only the second and third step are required, which is handled in Table~\ref{tbl:measured_time_to_query_the_datasets_warm}.
Here, \texttt{wget} was used to perform queries to the loaded web server.

\begin{table}[htb!]
\centering
\caption{Measured time to query the datasets -- cold; values rounded to 3 decimals.}
\label{tbl:measured_time_to_query_the_datasets_cold}
\begin{tabular}{lrrrr}
\toprule
dataset                  & json index size [MB] & npz index size [MB] & average time [s] & std time [s]  \\
\midrule
\texttt{avt\_ai\_images} & 1.6                  & 0.6                 & 3.763            & 0.093 \\
\texttt{sample200}       & 2.2                  & 0.8                 & 3.787            & 0.066 \\
\texttt{div2k}           & 9.8                  & 3.6                 & 3.802            & 0.071 \\
\texttt{sophoappeal}     & 11.5                 & 4.6                 & 3.823            & 0.088 \\
\texttt{avt\_image\_db}  & 12.3                 & 4.6                 & 3.954            & 0.603 \\
\texttt{unsplash-lite}   & 271.3                & 101.2               & 3.998            & 0.178 \\
\bottomrule
\end{tabular}
\end{table}

\begin{table}[htb!]
\centering
\caption{Measured time to query the datasets -- warm; values rounded to 3 decimals.}
\label{tbl:measured_time_to_query_the_datasets_warm}
\begin{tabular}{lrr}
\toprule
dataset                  & average time [s] & std time [s]  \\
\midrule
\texttt{avt\_ai\_images} & 0.039            & 0.011 \\
\texttt{sample200}       & 0.039            & 0.009 \\
\texttt{div2k}           & 0.039            & 0.012 \\
\texttt{sophoappeal}     & 0.040            & 0.011 \\
\texttt{avt\_image\_db}  & 0.039            & 0.011 \\
\texttt{unsplash-lite}   & 0.057            & 0.012 \\
\bottomrule
\end{tabular}
\end{table}

Comparing Tables~\ref{tbl:measured_time_to_query_the_datasets_cold}, and~\ref{tbl:measured_time_to_query_the_datasets_warm}, it can be concluded that the majority of boot up time is spent for the loading of the index, even though this is nearly independent of the number of images in the index.
Furthermore, the time to process one query in a warm start simulation, indicates a fast processing of max $57$\,ms for \texttt{unsplash-lite}.
Also, the processing of the query seems to be independent of the number of images in the index.

\section{Conclusion}
In this paper a brief overview of CLIPSE, an Open-CLIP-based image search engine, was provided.
Various possible open-source search engines are existing; however, the majority of them are hard to set up and may be better suitable for larger datasets.
CLIPSE aims to be minimalistic considering the requirements for setup and is focused on small datasets for, e.g., specific research use cases.
To implement this simplified approach, Python, UV and modern HTML are used for the system.
Furthermore, a brief evaluation considering the time needed to index and query is provided.
Several selected datasets are used and compared for both scenarios, considering the time and size of the datasets.
It is shown that for smaller datasets, there is no direct dependency on the number of images included in the index.
In the future, CLIPSE could also be applied in a distributed approach to handle even larger datasets.

\bibliographystyle{apalike}
\bibliography{refs}

\end{document}